\documentclass[11pt]{article}

\title{Joint and conditional estimation of tagging and parsing models\thanks{
 \noindent\, I would like to thank Eugene Charniak and the other
 members of BLLIP for their comments and suggestions.  Fernando
 Pereira was especially generous with comments and suggestions, as
 were the ACL reviewers; I apologize for not being able to follow up
 all of your good suggestions.  This research was supported by NSF
 awards 9720368 and 9721276 and NIH award R01 MH60922-01A2.}}
\author{Mark Johnson
\\ Brown University \\ {\small\sf Mark\_Johnson@Brown.edu}}

\usepackage{graphics}
\usepackage{color}
\usepackage{xy}
\xyoption{all}
\xyoption{dvips}

\usepackage{acl2001,times}
\renewcommand{\Pr}{{\rm P}}
\newcommand{\Prhat}{\hat{\Pr}}
\newcommand{\E}{{\rm E}}
\newcommand{\argmax}{\mathop{\rm argmax}}
\newcommand{\ra}{\raise 0.75pt \hbox{$\scriptscriptstyle\rightarrow$}}
\newcommand{\st}{:}

\newcommand{\thetaj}{\hat{\theta}}
\newcommand{\thetac}{\hat{\theta}}

\newcommand{\LD}{\Pr(\vec{y})}
\newcommand{\CLD}{\Pr(\vec{y}|\vec{x})}
\newcommand{\LX}{\Pr(\vec{x})}

\newcommand{\EOS}{\star}
\newcommand{\shift}{{\rm shift}}
\newcommand{\reduce}{{\rm reduce}}

\newcommand{\namecite}[1]{\newcite{#1}}

\begin{document}

\maketitle

\begin{abstract}
This paper compares two different ways of estimating statistical
language models.  Many statistical NLP tagging and parsing models are
estimated by maximizing the (joint) likelihood of the fully-observed
training data.  However, since these applications only require the
conditional probability distributions, these distributions can in
principle be learnt by maximizing the conditional likelihood of the
training data.  Perhaps somewhat surprisingly, models
estimated by maximizing the joint were superior to models estimated
by maximizing the conditional, even though some of the latter models intuitively
had access to ``more information''.
\end{abstract}

\section{Introduction}
Many statistical NLP applications, such as tagging and parsing,
involve finding the value of some hidden variable $Y$ (e.g., a tag or
a parse tree) which maximizes a conditional probability distribution
$\Pr_\theta(Y|X)$, where $X$ is a given word string.  The model
parameters $\theta$ are typically estimated by maximum likelihood:
i.e., maximizing the likelihood of the training data.  Given a (fully
observed) training corpus $D = ((y_1,x_1), \ldots, (y_n,x_n))$, the
{\em maximum (joint) likelihood estimate} (MLE) of $\theta$ is:
\begin{eqnarray}
\thetaj & = &
  \argmax_\theta \, \prod_{i=1}^n \Pr_\theta(y_i,x_i). \label{e:mle}
\end{eqnarray}

However, it turns out there is another maximum likelihood estimation
method which maximizes the conditional likelihood or ``pseudo-likelihood''
of the training data \cite{Besag75}.  Maximum conditional likelihood
is consistent {\em for the conditional distribution}.  Given a
training corpus $D$, the {\em maximum conditional likelihood estimate} (MCLE)
of the model parameters $\theta$ is:
\begin{eqnarray}
\thetac & = &
  \argmax_\theta \, \prod_{i=1}^n \Pr_\theta(y_i|x_i). \label{e:mcle}
\end{eqnarray}

\begin{figure}
\input{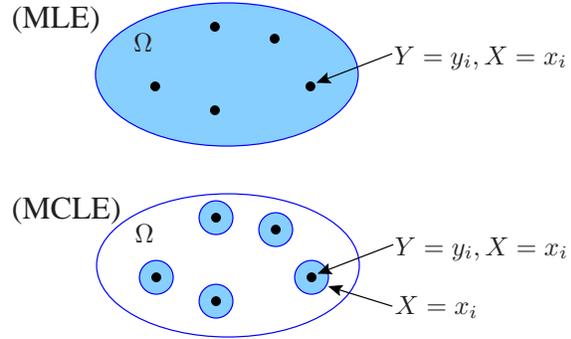}
\caption{\small The MLE makes the training data $(y_i,x_i)$ as likely
as possible (relative to $\Omega$), while the MCLE makes 
$(y_i,x_i)$ as likely as possible relative to other pairs $(y',x_i)$.
\label{f:mle}}
\end{figure}

Figure~\ref{f:mle} graphically depicts the difference between the MLE
and MCLE.  Let $\Omega$ be the universe of all possible pairs $(y,x)$
of hidden and visible values.  Informally, the MLE selects the model
parameter $\theta$ which make the training data pairs $(y_i,x_i)$ as
likely as possible relative to all other pairs $(y',x')$ in $\Omega$.
The MCLE, on the other hand, selects the model parameter $\theta$ in
order to make the training data pair $(y_i,x_i)$ more likely than
other pairs $(y',x_i)$ in $\Omega$, i.e., pairs with the same visible 
value $x_i$ as the training datum.

In statistical computational linguistics, maximum conditional
likelihood estimators have mostly been used with general exponential
or ``maximum entropy'' models because standard maximum likelihood
estimation is usually computationally intractable
\cite{Berger96,DellaPietra97,Jelinek97}. 
Well-known computational linguistic models such as Maximum-Entropy
Markov Models \cite{McCallum00} and Stochastic Unification-based
Grammars \cite{Johnson99c} are standardly estimated with conditional
estimators, and it would be interesting to know whether conditional
estimation affects the quality of the estimated model.  It should be
noted that in practice, the MCLE of a model with a large number of
features with complex dependencies may yield far better performance
than the MLE of the much smaller model that could be estimated with
the same computational effort.  Nevertheless, as this paper shows,
conditional estimators can be used with other kinds of models besides
MaxEnt models, and in any event it is interesting to ask whether the
MLE differs from the MCLE in actual applications, and if so, how.

Because the MLE is consistent for the joint distribution $\Pr(Y,X)$
(e.g., in a tagging application, the distribution of word-tag
sequences), it is also consistent for the conditional distribution
$\Pr(Y|X)$ (e.g., the distribution of tag sequences given word
sequences) and the marginal distribution $\Pr(X)$ (e.g., the
distribution of word strings).  On the other hand, the MCLE 
is consistent for the conditional distribution $\Pr(Y|X)$ alone,
and provides no information about either the joint or the marginal
distributions.  Applications such as language modelling for speech
recognition and EM procedures for estimating from hidden data either
explicitly or implicitly require marginal distributions over the
visible data (i.e., word strings), so it is not statistically 
sound to use MCLEs for such applications.  On the other hand,
applications which involve predicting the value of the hidden
variable from the visible variable (such as tagging or parsing)
usually only involve the conditional distribution, which the
MCLE estimates directly.

Since both the MLE and MCLE are consistent for the conditional
distribution, both converge {\em in the limit} to the ``true''
distribution {\em if the true distribution is in the model class}.
However, given that we often have insufficient data in computational
linguistics, and there are good reasons to believe that the true
distribution of sentences or parses cannot be described by our models,
there is no reason to expect these asymptotic results to hold in
practice, and in the experiments reported below the MLE and MCLE
behave differently experimentally.

A priori, one can advance plausible arguments in favour of both the
MLE and the MCLE.  Informally, the MLE and the MCLE differ in the
following way.  Since the MLE is obtained by maximizing
$\prod_{i}\Pr_\theta(y_i|x_i)\Pr_\theta(x_i)$, the MLE exploits
information about the distribution of word strings $x_i$ in the
training data that the MCLE does not.  Thus one might expect the MLE
to converge faster than the MCLE in situations where training data is
not over-abundant, which is often the case in computational
linguistics.

On the other hand, since the intended application requires a
conditional distribution, it seems reasonable to directly estimate
this conditional distribution from the training data as the MCLE does.
Furthermore, suppose that the model class is wrong (as is surely true
of all our current language models), i.e., the ``true'' model
$\Pr(Y,X)\neq \Pr_\theta(Y,X)$ for all $\theta$, and that our best
models are particularly poor approximations to the true distribution of word
strings $\Pr(X)$.  Then ignoring the distribution of word strings in
the training data as the MCLE does might indeed be a reasonable thing
to do.

The rest of this paper is structured as follows.  The next section
formulates the MCLEs for HMMs and PCFGs as constrained optimization
problems and describes an iterative dynamic-programming method for
solving them.  Because of the computational complexity of these problems,
the method is only applied to a simple PCFG based on the ATIS corpus.
For this example, the MCLE PCFG does perhaps produce slightly better
parsing results than the standard MLE (relative-frequency) PCFG, although
the result does not reach statistical significance.

It seems to be difficult to find model classes for which the MLE and
MCLE are both easy to compute.  However, often it is possible to find
two closely related model classes, one of which has an easily computed
MLE and the other which has an easily computed MCLE.  Typically,
the model classes which have an easily computed MLE define {\em joint}
probability distributions over both the hidden and the visible data
(e.g., over word-tag pair sequences for tagging), while the model
classes which have an easily computed MCLE define {\em conditional}
probability distributions over the hidden data given the visible
data (e.g., over tag sequences given word sequences).

Section~\ref{s:tagging} investigates closely related joint and
conditional tagging models (the latter can be regarded as a
simplification of the Maximum Entropy Markov Models of
\namecite{McCallum00}), and shows that MLEs outperform the MCLEs in
this application.  The final empirical section investigates two
different kinds of stochastic shift-reduce parsers, and shows that the
model estimated by the MLE outperforms the model estimated by the
MCLE.

\section{PCFG parsing}
In this application, the pairs $(y,x)$ consist of a parse tree $y$ and
its terminal string or yield $x$ (it may be simpler to think of $y$ containing
all of the parse tree except for the string $x$).  Recall that in a PCFG with
production set $R$, each production $(A \ra \alpha) \in R$ is
associated with a parameter $\theta_{A \ra \alpha}$.  These parameters
satisfy a normalization constraint for each nonterminal $A$:
\begin{eqnarray}
\sum_{\alpha \st (A\ra\alpha)\in R}  \theta_{A\ra\alpha} & = & 1 \label{e:pcfgc}
\end{eqnarray}
For each production $r\in R$, let $f_r(y)$ be the number of times $r$ is
used in the derivation of the tree $y$.  Then the PCFG defines a probability
distribution over trees:
\begin{eqnarray*}
\Pr_\theta(Y) & = & \prod_{(A\ra\alpha)\in R} {\theta_{A\ra\alpha}}^{f_{A\ra\alpha}(Y)}
\end{eqnarray*}

The MLE for $\theta$ is the well-known ``relative-frequency'' estimator:
\begin{eqnarray*}
\thetaj_{A\ra\alpha} & = & { \sum_{i=1}^n f_{A\ra\alpha}(y_i) \over
  \sum_{i=1}^n \sum_{\alpha'\st (A\ra\alpha')\in R} f_{A\ra\alpha'}(y_i)}.
\end{eqnarray*}

Unfortunately the MCLE for a PCFG is more complicated.
If $x$ is a word string, then let $\tau(x)$ be the set of parse trees
with terminal string or yield $x$ generated by the PCFG.  Then given a
training corpus $D = ((y_1,x_1), \ldots, (y_n,x_n))$, where $y_i$ is a
parse tree for the string $x_i$, the log conditional likelihood of the
training data $\log \CLD$ and
its derivative are given by:
\begingroup
\small
\begin{eqnarray*}
\hspace*{-1em}
\log \CLD & = & \sum_{i=1}^n \left(\log\Pr_\theta(y_i) - 
       \log\sum_{y\in\tau(x_i)}\Pr_\theta(y)\right)  \label{e:pcfgcl} \\
\hspace*{-0.25em}
{\partial \log \CLD \over \partial \theta_{A\ra\alpha}}
 &=&
 {1 \over \theta_{A\ra\alpha}} \sum_{i=1}^n \left( f_{A\ra\alpha}(y_i) -  
  \E_\theta(f_{A\ra\alpha}|x_i) \right)
\end{eqnarray*}
\endgroup
Here $\E_\theta(f|x)$ denotes the expectation of $f$ with respect to
$\Pr_\theta$ conditioned on $Y\in\tau(x)$.  There does not seem to be
a closed-form solution for the $\theta$ that maximizes
$\CLD$ subject to the constraints (\ref{e:pcfgc}), so we
used an iterative numerical gradient ascent method, with the
constraints (\ref{e:pcfgc}) imposed at each iteration using Lagrange
multipliers.  Note that $\sum_{i=1}^n \E_\theta(f_{A\ra\alpha}|x_i)$
is a quantity calculated in the Inside-Outside algorithm \cite{Lari90}
and $\CLD$ is easily computed as a by-product of the same dynamic
programming calculation.

Since the expected production counts $\E_\theta(f|x)$ depend on the
production weights $\theta$, the entire training corpus must be reparsed
on each iteration (as is true of the Inside-Outside algorithm).  This
is computationally expensive with a large grammar and training corpus;
for this reason the MCLE PCFG experiments described here were performed
with the relatively small ATIS treebank corpus of air travel reservations
distributed by LDC.

In this experiment, the PCFGs were always trained on the 1088
sentences of the ATIS1 corpus and evaluated on the 294 sentences of
the ATIS2 corpus.  Lexical items were ignored; the PCFGs generate
preterminal strings.  The iterative algorithm for the MCLE was
initialized with the MLE parameters, i.e., the ``standard'' PCFG
estimated from a treebank.  Table~\ref{t:pcfg} compares the MLE and
MCLE PCFGs.

\begin{table}
\begin{center}
\begin{tabular}{lrr}
               & MLE     & MCLE \\
$-\log \LD$   &  13857   & 13896 \\
$-\log \CLD$  &  1833    & 1769  \\
$-\log \LX$   &  12025   & 12127 \\
Labelled precision & 0.815 & 0.817 \\
Labelled recall    & 0.789 & 0.794
\end{tabular}
\end{center}
\caption{\label{t:pcfg} \small
 The likelihood $\LD$ and conditional likelihood $\CLD$ of the ATIS1 training trees,
 and the marginal likelihood $\LX$ of the ATIS1 training strings, as well as the labelled
 precision and recall of the ATIS2 test trees, using the MLE and MCLE PCFGs.}
\end{table}

The data in table~\ref{t:pcfg} shows that compared to the MLE PCFG,
the MCLE PCFG assigns a higher conditional probability of the parses
in the training data given their yields, at the expense of assigning a
lower marginal probability to the yields themselves.  The labelled
precision and recall parsing results for the MCLE PCFG were slightly
higher than those of the MLE PCFG.  Because both the test data set and
the differences are so small, the significance of these results was
estimated using a bootstrap method with the difference in F-score in
precision and recall as the test statistic \cite{Cohen95}.  This test
showed that the difference was not significant ($p \approx 0.1$). 
Thus the MCLE PCFG did not perform significantly better than the MLE
PCFG in terms of precision and recall.

\section{HMM tagging} \label{s:tagging}
As noted in the previous section, maximizing the conditional
likelihood of a PCFG or a HMM can be computationally intensive.  This
section and the next pursues an alternative strategy for comparing
MLEs and MCLEs: we compare similiar (but not identical) model classes,
one of which has an easily computed MLE, and the other of which has an
easily computed MCLE.  The application considered in this section is
bitag POS tagging, but the techniques extend straight-forwardly to
$n$-tag tagging.  In this application, the data pairs $(y,x)$ consist
of a tag sequence $y = t_1 \ldots t_m$ and a word sequence $x = w_1
\ldots w_m$, where $t_j$ is the tag for word $w_j$ (to simplify the
formulae, $w_0$, $t_0$, $w_{m+1}$ and $t_{m+1}$ are always taken to be
end-markers).  Standard HMM tagging models define a {\em joint}
distribution over word-tag sequence pairs; these are most
straight-forwardly estimated by maximizing the likelihood of the joint
training distribution.  However, it is straight-forward to devise
closely related HMM tagging models which define a {\em conditional}
distribution over tag sequences given word sequences, and which are
most straight-forwardly estimated by maximizing the conditional
likelihood of the distribution of tag sequences given word sequences
in the training data.  

\begin{figure}
\vspace*{-0.24in}
\entrymodifiers={++[o][F-]}
\newcommand{\mybox}[1]{\hbox to 2.25em {\hfill$#1$\hfill}}
\xymatrix{ 
*{(\ref{e:hmm})} & *{\;\cdots\;}\ar[r] & \mybox{T_j} \ar[r]\ar[d] & \mybox{T_{j+1}} \ar[r]\ar[d] & *{\;\cdots\;} \\
*{}              & *{}             & \mybox{W_j}              & \mybox{W_{j+1}}             
}
\vspace{1em}
\xymatrix{
*{(\ref{e:chmm})} & *{\;\cdots\;}\ar[r] & \mybox{T_j} \ar[r] & \mybox{T_{j+1}} \ar[r] & *{\;\cdots\;} \\
*{}               & *{}       & \mybox{W_j} \ar[u] & \mybox{W_{j+1}} \ar[u]
}
\vspace{1em}
\xymatrix{
*{(\ref{e:hmm1})} & *{\;\cdots\;}\ar[r] & \mybox{T_j} \ar[r]\ar[d] & \mybox{T_{j+1}} \ar[r]\ar[d] & *{\;\cdots\;} \\
*{}               & *{}\ar[ur]& \mybox{W_j} \ar[ur]      & \mybox{W_{j+1}} \ar[ur]
}
\vspace{1em}
\xymatrix{
*{(\ref{e:hmm2})} & *{\mybox{\cdots}}\ar[r]\ar[dr] & \mybox{T_j} \ar[r]\ar[dr] & \mybox{T_{j+1}} \ar[r]\ar[dr] & *{\;\cdots\;} \\
*{}               & *{}       & \mybox{W_j} \ar[u] & \mybox{W_{j+1}} \ar[u] & *{}
}
\caption{\small The HMMs depicted as ``Bayes net'' graphical models. \label{f:bayes}}
\end{figure}
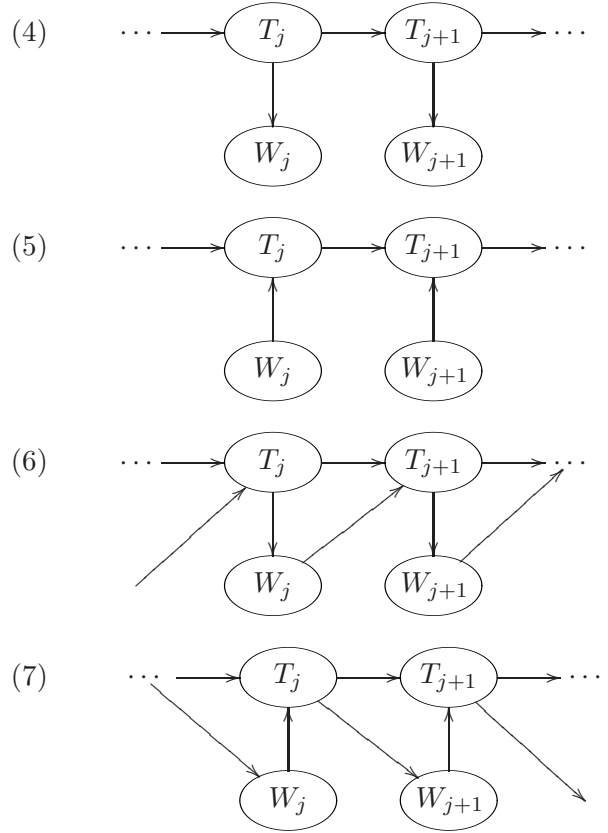

All of the HMM models investigated in this
section are instances of a certain kind of graphical model that
\namecite{Pearl88} calls ``Bayes nets''; Figure~\ref{f:bayes}
sketches the networks that correspond to all of the models discussed
here.  (In such a graph, the set of incoming arcs to a node depicting
a variable indicate the set of variables on which this variable is
conditioned).

Recall the standard bitag HMM model, which defines a joint distribution
over word and tag sequences:
\begin{eqnarray}
\Pr(Y,X) &=& \prod_{j=1}^{m+1} \Prhat(T_j|T_{j-1}) \Prhat(W_j|T_j) 
    \label{e:hmm}
\end{eqnarray}
As is well-known, the MLE for (\ref{e:hmm}) sets $\Prhat$ to the
empirical distributions on the training data.

Now consider the following {\em conditional model} of the conditional
distribution of tags given words (this is a simplified form of the
model described in \namecite{McCallum00}):
\begin{eqnarray}
\Pr(Y|X) &=& \prod_{j=1}^{m+1} \Pr_0(T_j|W_j,T_{j-1}) \label{e:chmm}
\end{eqnarray}
The MCLE of (\ref{e:chmm}) is easily calculated: $\Pr_0$ should be set
the empirical distribution of the training data.  However, to minimize
sparse data problems we estimated $\Pr_0(T_j|W_j,T_{j-1})$ as a
mixture of $\Prhat(T_j|W_j)$, $\Prhat(T_j|T_{j-1})$ and
$\Prhat(T_j|W_j,T_{j-1})$, where the $\Prhat$ are empirical
probabilities and the (bucketted) mixing parameters are determined
using deleted interpolation from heldout data \cite{Jelinek97}.

These models were trained on sections 2-21 of the Penn tree-bank
corpus.  Section 22 was used as heldout data to evaluate the
interpolation parameters $\lambda$.  The tagging accuracy of the
models was evaluated on section 23 of the tree-bank corpus (in both
cases, the tag $t_j$ assigned to word $w_j$ is the one which maximizes
the marginal $\Pr(t_j|w_1 \ldots w_m)$, since this minimizes the
expected loss on a tag-by-tag basis).  

The conditional model (\ref{e:chmm}) has the worst performance of any
of the tagging models investigated in this section: its tagging
accuracy is 94.4\%.  The joint model (\ref{e:hmm}) has a considerably
lower error rate: its tagging accuracy is 95.5\%.

One possible explanation for this result is that the way in which the
interpolated estimate of $\Pr_0$ is calculated, rather than
conditional likelihood estimation per se, is lowering tagger accuracy
somehow.  To investigate this possibility, two additional joint models
were estimated and tested, based on the formulae below.
\begingroup
\begin{eqnarray}
\hspace*{-2.9em}
\Pr(Y,X) &\!\!=\!\!& \prod_{j=1}^{m+1} \Prhat(W_j|T_j) \Pr_1(T_j|W_{j-1},T_{j-1}) \label{e:hmm1} \\
\hspace*{-2.9em}
\Pr(Y,X) &\!\!=\!\!& \prod_{j=1}^{m+1} \Pr_0(T_j|W_j,T_{j-1}) \Prhat(W_j|T_{j-1}) \label{e:hmm2}
\end{eqnarray}
\endgroup
The MLEs for both (\ref{e:hmm1}) and (\ref{e:hmm2}) are easy to
calculate.  (\ref{e:hmm1}) contains a conditional distribution $\Pr_1$
which would seem to be of roughly equal complexity to $\Pr_0$, and it
was estimated using deleted interpolation in exactly the same way as
$\Pr_0$, so if the poor performance of the conditional model was due
to some artifact of the interpolation procedure, we would expect the
model based on (\ref{e:hmm1}) to perform poorly.  Yet the tagger based
on (\ref{e:hmm1}) performs the best of all the taggers investigated in
this section: its tagging accuracy is 96.2\%.

(\ref{e:hmm2}) is admitted a rather strange model, since the right
hand term in effect predicts the {\em following} word from the current
word's tag.  However, note that (\ref{e:hmm2}) differs from
(\ref{e:chmm}) only via the presence of this rather unusual term,
which effectively converts (\ref{e:chmm}) from a conditional model to
a joint model.  Yet adding this term improves tagging accuracy considerably,
to 95.3\%.  Thus for bitag tagging at least, the conditional model has
a considerably higher error rate than any of the joint models examined
here.  (While a test of significance was not conducted here, previous
experience with this test set shows that performance differences of this
magnitude are extremely significant statistically).

\section{Shift-reduce parsing}
The previous section compared similiar joint and conditional tagging
models.  This section compares a pair of joint and conditional parsing
models.  The models are both stochastic shift-reduce parsers; they
differ only in how the distribution over possible next moves are
calculated.  These parsers are direct simplifications of the
Structured Language Model \cite{Jelinek00}.  Because the parsers'
moves are determined solely by the top two category labels on the
stack and possibly the look-ahead symbol, they are much simpler than
stochastic LR parsers \cite{Briscoe93,Inui97}.  The distribution over
trees generated by the joint model is a probabilistic context-free language
\cite{Abney99a}.
As with the PCFG models discussed earlier, these parsers are not
lexicalized; lexical items are ignored, and the POS tags are used as
the terminals.

\begin{figure}
\begin{center}
\input{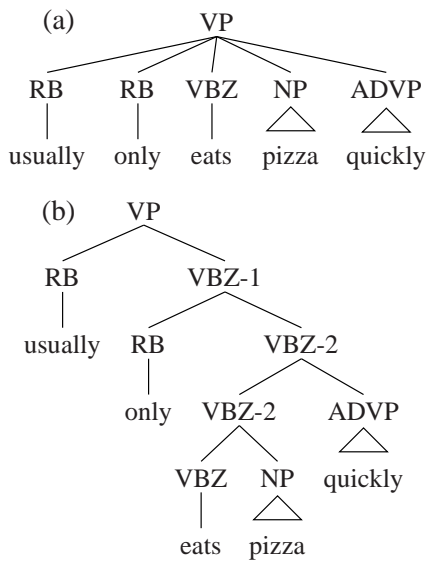}
\end{center}
\caption{\small The binarization transformation used in the shift-reduce
 parser experiments transforms tree (a) into tree (b). \label{f:bin}}
\end{figure}

These two parsers only produce trees with unary or binary nodes, so we
binarized the training data before training the parser, and debinarize
the trees the parsers produce before evaluating them with respect to
the test data \cite{Johnson98c}.  We binarized by inserting $n-2$
additional nodes into each local tree with $n > 2$ children.  We
binarized by first joining the head to all of the constituents to its
right, and then joining the resulting structure with constituents to
the left.  The label of a new node is the label of the head followed
by the suffix ``-1'' if the head is (contained in) the right child or
``-2'' if the head is (contained in) the left child.  Figure~\ref{f:bin}
depicts an example of this transformation.

The Structured Language Model is described in detail in
\namecite{Jelinek00}, so it is only reviewed here.  Each parser's
stack is a sequence of node labels (possibly including labels
introduced by binarization).  In what follows, $s_1$ refers to the top
element of the stack, or `$\EOS$' if the stack is empty; similarly
$s_2$ refers to the next-to-top element of the stack or `$\EOS$' if
the stack contains less than two elements.  We also append a `$\EOS$'
to end of the actual terminal string being parsed (just as with the
HMMs above), as this simplifies the formulation of the parsers, i.e.,
if the string to be parsed is $w_1 \ldots w_m$, then we take $w_{m+1}
= \EOS$.

A shift-reduce parse is defined in terms of moves.  A move is either
$\shift(w)$, $\reduce_1(c)$ or $\reduce_2(c)$, where $c$ is a
nonterminal label and $w$ is either a terminal label or `$\EOS$'.
Moves are partial functions from stacks to stacks: a $\shift(w)$ move
pushes a $w$ onto the top of stack, while a $\reduce_i(c)$ move pops
the top $i$ terminal or nonterminal labels off the stack and pushes a
$c$ onto the stack.  A shift-reduce parse is a sequence of moves which
(when composed) map the empty stack to the two-element stack whose top
element is `$\EOS$' and whose next-to-top element is the start symbol.
(Note that the last move in a shift-reduce parse must always be a
$\shift(\EOS)$ move; this corresponds to the final ``accept'' move in
an LR parser).  The isomorphism between shift-reduce parses and
standard parse trees is well-known \cite{Hopcroft79}, and so is not
described here.

A (joint) shift-reduce parser is defined by a distribution
$\Pr(m|s_1,s_2)$ over next moves $m$ given the top and next-to-top
stack labels $s_1$ and $s_2$.  To ensure that the next move is in fact
a possible move given the current stack, we require that
$\Pr(\reduce_1(c)|\EOS,\EOS) = 0$ and $\Pr(\reduce_2(c)|c',\EOS) = 0$
for all $c, c'$, and that $\Pr(shift(\EOS)|s_1,s_2) = 0$ unless $s_1$
is the start symbol and $s_2 = \EOS$.  Note that this extends to a
probability distribution over shift-reduce parses (and hence parse
trees) in a particularly simple way: the probability of a parse is the
product of the probabilities of the moves it consists of.  Assuming
that $\Pr$ meets certain tightness conditions, this distribution over
parses is properly normalized because there are no ``dead'' stack
configurations: we require that the distribution over moves be defined
for all possible stacks.

A conditional shift-reduce parser differs only minimally from the
shift-reduce parser just described: it is defined by a distribution
$\Pr(m|s_1,s_2,t)$ over next moves $m$ given the top and next-to-top
stack labels $s_1$, $s_2$ and the next input symbol $w$ ($w$ is called
the {\em look-ahead symbol}).  In addition to the requirements on
$\Pr$ above, we also require that if $w' \neq w$ then $\Pr(\shift(w')
| s_1,s_2,w) = 0$ for all $s_1, s_2$; i.e., shift moves can only shift
the current look-ahead symbol.  This restriction implies that all
non-zero probability derivations are derivations of the parse string,
since the parse string forces a single sequence of symbols to be
shifted in all derivations.  As before, since there are no ``dead''
stack configurations, so long as $\Pr$ obeys certain tightness
conditions, this defines a properly normalized distribution over parses.
Since all the parses are required to be parses of of the input
string, this defines a conditional distribution over parses given the
input string.

It is easy to show that the MLE for the joint model, and the MCLE for
the conditional model, are just the empirical distributions from the
training data.  We ran into sparse data problems using the empirical
training distribution as an estimate for $\Pr(m|s_1,s_2,w)$ in the
conditional model, so in fact we used deleted interpolation to
interpolate $\Prhat(m|s_1,s_2,w)$, and $\Prhat(m|s_1,s_2)$ to estimate
$\Pr(m|s_1,s_2,w)$.  The models were estimated from sections 2--21 of
the Penn treebank, and tested on the 2245 sentences of length~40 or
less in section 23.  The deleted interpolation parameters were
estimated using heldout training data from section 22.

We calculated the most probable parses using a dynamic programming
algorithm based on the one described in \namecite{Jelinek00}.  Jelinek
notes that this algorithm's running time is $n^6$ (where $n$ is the
length of sentence being parsed), and we found exhaustive parsing to
be computationally impractical.  We used a beam search procedure which
thresholded the best analyses of each prefix of the string being
parsed, and only considered analyses whose top two stack symbols had
been observed in the training data.  In order to help guard against
the possibility that this stochastic pruning influenced the results,
we ran the parsers twice, once with a beam threshold of $10^{-6}$
(i.e., edges whose probability was less than $10^{-6}$ of the best
edge spanning the same prefix were pruned) and again with a beam
threshold of $10^{-9}$.  The results of the latter runs are reported
in table~\ref{t:sr}; the labelled precision and recall results from
the run with the more restrictive beam threshold differ by less than
$0.001$, i.e., at the level of precision reported here,
are identical with the results presented in table~\ref{t:sr}
except for the Precision of the Joint SR parser, which was $0.665$.  
For comparision, table~\ref{t:sr} also reports results from
the non-lexicalized treebank PCFG estimated from the transformed trees
in sections 2-21 of the treebank; here exhaustive CKY parsing was used
to find the most probable parses.

\begin{table}
\begin{center}
\hspace*{-2em}
\begin{tabular}{lccc}
          & Joint SR & Conditional SR & PCFG \\
Precision & 0.666    &  0.633         & 0.700 \\
Recall    & 0.650    &  0.639         & 0.657     
\end{tabular}
\end{center}
\caption{\small Labelled precision and recall results for joint and conditional shift-reduce parsers,
         and for a PCFG. \label{t:sr}}
\end{table}

All of the precision and recall results, including those for the PCFG,
presented in table~\ref{t:sr} are much lower than those from
a standard treebank PCFG; presumably this is because the binarization
transformation depicted in Figure~\ref{f:bin} loses information about
pairs of non-head constituents in the same local tree
(\namecite{Johnson98c} reports similiar performance degradation for
other binarization transformations).  Both the joint and the
conditional shift-reduce parsers performed much worse than
the PCFG.  This may be due to the pruning effect of the beam search,
although this seems unlikely given that varying the beam threshold did
not affect the results.  The performance difference between the joint
and conditional shift-reduce parsers bears directly on the issue
addressed by this paper: the joint shift-reduce parser performed much
better than the conditional shift-reduce parser.  The differences are
around a percentage point, which is quite large in parsing research
(and certainly highly significant).

The fact that the joint shift-reduce parser outperforms the
conditional shift-reduce parser is somewhat surprising.  Because the
conditional parser predicts its next move on the basis of the
lookahead symbol as well as the two top stack categories, one might
expect it to predict this next move more accurately than the joint
shift-reduce parser.  The results presented here show that this is not
the case, at least for non-lexicalized parsing.  The {\em label bias}
of conditional models may be responsible for this \cite{Bottou91,Lafferty01}.

\section{Conclusion}
This paper has investigated the difference between maximum likelihood
estimation and maximum conditional likelihood estimation for three
different kinds of models: PCFG parsers, HMM taggers and shift-reduce
parsers.  The results for the PCFG parsers suggested that conditional
estimation might provide a slight performance improvement, although
the results were not statistically significant since computational
difficulty of conditional estimation of a PCFG made it necessary to
perform the experiment on a tiny training and test corpus.  In order
to avoid the computational difficulty of conditional estimation, we
compared closely related (but not identical) HMM tagging and
shift-reduce parsing models, for some of which the maximum likelihood
estimates were easy to compute and for others of which the maximum
conditional likelihood estimates could be easily computed.  In both
cases, the joint models outperformed the conditional models by quite
large amounts.  This suggests that it may be worthwhile investigating
methods for maximum (joint) likelihood estimation for model classes
for which only maximum conditional likelihood estimators are currently
used, such as Maximum Entropy models and MEMMs, since if the results
of the experiments presented in this paper extend to these models,
one might expect a modest performance improvement.

As explained in the introduction, because maximum likelihood
estimation exploits not just the conditional distribution of hidden
variable (e.g., the tags or the parse) conditioned on the visible
variable (the terminal string) but also the marginal distribution of
the visible variable, it is reasonable to expect that it should
outperform maximum conditional likelihood estimation.  Yet it is
counter-intuitive that joint tagging and shift-reduce parsing models,
which predict the next tag or parsing move on the basis of what seems
to be less information than the corresponding conditional model,
should nevertheless outperform that conditional model, as the
experimental results presented here show.  The recent theoretical and
simulation results of \namecite{Lafferty01} suggest that conditional
models may suffer from {\em label bias} (the discovery of which Lafferty et. al.
attribute to \namecite{Bottou91}), which may provide an insightful explanation
of these results.

None of the models investigated here are state-of-the-art; the goal
here is to compare two different estimation procedures, and for that
reason this paper concentrated on simple, easily implemented models.
However, it would also be interesting to compare the performance of
joint and conditional estimators on more sophisticated models.  

\bibliographystyle{acl}
\bibliography{mj}

\end{document}